\documentclass{article}[draft]
\usepackage{spconf,amsmath,graphicx}

\usepackage{enumitem}
\setlist{nosep, leftmargin=14pt}

\usepackage{mwe} 

\usepackage{pifont}
\newcommand{\cmark}{\ding{51}}
\newcommand{\xmark}{\ding{55}}

\usepackage{xspace}  

\def\vitamino{ViTAMIn-O\xspace}

\newcommand{\vect}[1]{\mathbf{#1}} 
\newcommand{\matr}[1]{\mathbf{#1}} 

\title{Interpretable deep concepts of organoid morphology}
\name{Lion J. Gleiter$^{\star \dagger}$ \qquad Tingying Peng$^{\star \dagger}$}

\address{
    $^{\star}$ Helmholtz Munich, Helmholtz AI, Neuherberg, Germany \\
    $^{\dagger}$ Technical University of Munich, School of CIT, Munich, Germany \\
}

\begin{document}
%
\maketitle
\begin{abstract}
Image-based classification of organoid morphologies can indicate drug response, viability, or successful differentiation. 
In recent work, learned representations outperform handcrafted features as the basis for downstream classifiers. 
However, their black-box nature hides the morphological cues a classifier relies on. 
Based on the foundation models DINOv3 (general-purpose) and \vitamino (organoid-specific), we train sparse autoencoders (SAEs) to learn interpretable concepts related to distinct organoid morphologies. 
We annotate concepts with spatial labels (organoid, organoid wall, background) and compute their dataset specificity and task relevance across five tasks. 
The spatial labels of DINOv3 concepts yield coarse organoid segmentations on out-of-domain datasets. 
Filtering concepts by spatial label in the DINOv3-SAE improves downstream AUROC for three of the five tasks. 
Qualitative analysis of the most relevant concepts further reveals a confounding factor introduced by our preprocessing in one task. 
Through such insights, interpretable concepts could help improve the reliability of classifiers and guide the future development of organoid representation models.
The trained SAEs will be made available upon publication.

\end{abstract}
\section{Introduction}
\label{sec:intro}

\begin{figure*}[htb]
    \centering
    \includegraphics[width=1.0\linewidth]{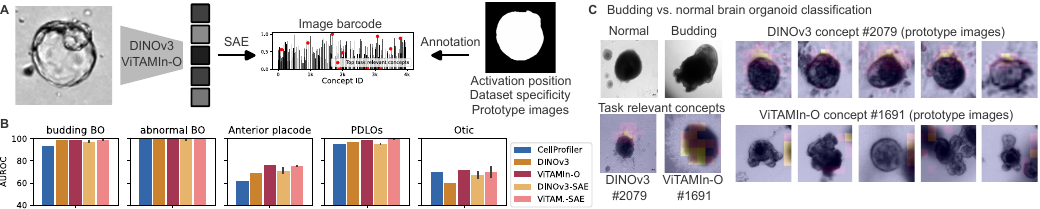}
    \caption{A) An SAE learns sparse concepts from dense foundation model representations. Activation position, dataset specificity, and reference images provide interpretable annotations per concept. 
    B) Foundation models outperform CellProfiler feature-based downstream classification in most cases. The corresponding SAE features only slightly reduce classification AUROC. C) Prototype images for the most task-relevant concepts for budding brain organoids appear to focus on texture around the organoid (DINOv3 \#2079) or a dark, textureless organoid border (ViTAMIn-O \#1691).}
    \label{fig:methods-overview}
\end{figure*}

Deep learning is increasingly used to process brightfield microscopy images of organoids~\cite{vitamino,OrganoID,OrgaSegment,OrgaQuant,Tellu,MorphoOrgaAgent}. After detection and segmentation, single organoids can for example be classified to predict drug response, differentiation, viability, or to select the optimal time point for further laboratory analysis. 
Deep learning models show superior performance compared to interpretable handcrafted morphometrics, such as shape, intensity, or texture-based features for organoid analysis and in other biomedical fields~\cite{vitamino,DINO4Cells}. However, training datasets in brightfield microscopy can exhibit batch effects or other confounding features.

For clinically relevant tasks but also to avoid false research conclusions and to verify hypotheses, it is important to understand the decision process of deep learning-based downstream predictors. 
Unlike CellProfiler features~\cite{CellProfiler}, which capture pre-defined shape, intensity, or texture-specific features, directions in deep feature spaces are difficult to assign to biological meaning. ACE~\cite{ACE} developed a framework to automatically generate concepts present in deep features by clustering representations from different image regions. \vitamino adapts this approach (OrgACE) to post-hoc annotate downstream decisions of a task-specific, finetuned model with patch-based concepts~\cite{vitamino}. 
Recent work investigates concepts in DINOv2 and their relevance for various natural image tasks~\cite{into-the-rabbit-hull} and disease-related concepts in a hematology foundation model~\cite{CytoSAE}. In both cases, concepts are learned with a Sparse Autoencoder (SAE), which extracts an over-complete dictionary of concepts from patch and class tokens of a deep representation model~\cite{into-the-rabbit-hull,CytoSAE}.


In this work, we study concepts present in a natural-image foundation model (DINOv3 large) and an organoid-specific generalist representation model (\vitamino) and their relevance for various organoid morphology-related tasks.
In contrast to OrgACE~\cite{vitamino}, we study organoid morphology-related concepts learned from generalist models instead of a single task-specific model, and evaluate on five diverse downstream tasks. 
Our contribution is two-fold: 
First, we train SAEs on frozen foundation models with a diverse organoid dataset to learn interpretable concepts of organoid morphology. 
We will make the code and weights publicly available upon publication such that our SAE concepts can be re-used without retraining. 
Second, we annotate the learned concepts with position labels (background, border, organoid), prototype images, dataset specificity, and task relevance. Unlike OrgACE, we annotate based on a large and diverse data pool from five different sources, which increases the diversity of potential morphologies and provides a metric of dataset specificity of concepts.

We show that the positional labeling of DINOv3 concepts enables a rough segmentation of background, border, and organoid on out-of-domain data. Further, we observe differences between DINOv3 and \vitamino features and observe that attribution-based filtering of relevant concepts can improve downstream classification for DINOv3 concepts. Qualitatively, we demonstrate that prototype images can hint at a confounding feature from preprocessing in one downstream task.

\section{Interpretable concept annotation for organoids}
\label{sec:interpretable-concept-annotation-organoids}

\subsection{Learning concepts}
\label{sec:interpretable-concept-annotation-organoids:learning-concepts}

SAEs are a common tool to extract interpretable concepts from the semantically entangled activations of a neural network~\cite{TopKSAE,BatchTopKSAE,RA_SAE,CytoSAE,into-the-rabbit-hull}. 
Similar to~\cite{into-the-rabbit-hull}, we train a BatchTopK sparse autoencoder~\cite{BatchTopKSAE} with a relaxed archetypal dictionary (RA-SAE)~\cite{RA_SAE} for stability.

A BatchTopK SAE~\cite{BatchTopKSAE} consists of an encoder $f(\vect{x}) = \operatorname{BatchTopK}(\matr{W}_{\text{enc}}\vect{x} + \vect{b}_{\text{enc}})$ and a decoder $g(\vect{z})=\matr{W}_{\text{dec}}\vect{z} + \vect{b}_{\text{dec}}$. The $\operatorname{BatchTopK}$ activation function sets all activations below a threshold to zero. During training, the threshold is chosen per batch such that all except the top K activations of the batch are zero. During inference, it uses an average of the thresholds tracked during training. BatchTopK SAEs directly control the level of sparsity through the ratio of K to the batch size and number of dictionary atoms~\cite{BatchTopKSAE}. The training loss is the same as for TopK SAEs~\cite{TopKSAE}, which complements the reconstruction MSE loss with an auxiliary loss $\mathcal{L}_{aux}$ to reduce the number of dead codes, i.e. codes that never activate. Since BatchTopK can distribute activations unequally across samples, inactive samples with zero active codes can occur during training. To incentivize at least one active code per sample, we add an auxiliary top-1 loss on inactive samples:
\begin{align}
    &\mathcal{L}(\vect{x}) = \| \vect{x} -  g ( f (\vect{x})) \|_2^2 + \lambda_{\text{aux}}\mathcal{L}_{\text{aux}}(\vect{x}) + \mathcal{L}_{\text{top1}}(\vect{x}_{\text{inact.}}) \\
    &\mathcal{L}_{\text{top1}}(\vect{x}) = \| \vect{x} -  g ( \operatorname{Top_1}(\matr{W}_{\text{enc}}\vect{x} + \vect{b}_{\text{enc}}) ) \|_2^2
\end{align}
Here, $\operatorname{Top_1}(\vect{x})$ sets all except the largest activation per sample to zero, $\mathcal{L}_{\text{aux}}$ is defined in~\cite{TopKSAE}, and $\lambda_{\text{aux}}=1/32$.

We replace the decoder $g$ with a relaxed archetypal (RA) dictionary~\cite{RA_SAE}. Instead of directly learning a decoder dictionary $\matr{W}_{\text{dec}}$, an archetypal dictionary precomputes a set of candidate dictionary atoms and constrains the decoder dictionary to convex combinations of the candidate points $\matr{C}$:
\begin{align}
    \matr{D}_{\text{ra}} = \matr{W}\matr{C} \quad \text{subject to }\matr{W}\text{ row-stochastic,}
\end{align}
i.e. $\matr{W}_{i,j}\geq 0$ and $\sum_{j} \matr{W}_{i,j} = 1$. To trade off the stability of a predefined dictionary space and the flexibility of a standard SAE, an RA dictionary adds relaxation weights $\matr{D}_{\text{ra}} = \matr{W}\matr{C} + \Lambda$ subject to a norm constraint $\| \Lambda \|_2^2 \leq \delta$~\cite{RA_SAE}. Instead of enforcing the row-stochasticity of $\matr{W}$ after each optimizer step, we add a row-wise $\operatorname{softmax}$ activation and leave $\matr{W}$ unconstrained. Our decoder function is therefore
\begin{align}
    g(\vect{z}) = (\operatorname{softmax}(\matr{W})\matr{C} + \Lambda) \vect{z}  \, .
\end{align}

\begin{table}[]
    \centering
    \caption{RA-SAE training configurations (* chosen value).}
    \label{table:ablation-sae-params}
    \small
    \begin{tabular}{c|c|c||c|c|c}
        \hline
        $\mathcal{L}_{\text{top1}}$ & $\operatorname{softmax}(\matr{W})$ & $\delta$ & $R^2$ & Stab. & \% dead \\
        \hline
        \multicolumn{3}{c||}{} & \multicolumn{3}{c}{\vitamino} \\
        \hline
        \xmark & \xmark & 0.1 & 0.62 & 0.51 & 55.7 \\
        \cmark & \xmark & 0.1 & 0.55 & 0.62 & 5.3 \\
        \cmark & \cmark & 0.1 & 0.56 & 0.55 & 1.6 \\
        \cmark & \cmark & 1.0 & 0.58* & 0.56* & 0.4* \\
        \hline
        \multicolumn{3}{c||}{} & \multicolumn{3}{c}{DINOv3} \\
        \hline
        \xmark & \xmark & 0.1 & 0.64 & 0.59 & 32.3 \\ 
        \cmark & \xmark & 0.1 & 0.63 & 0.59 & 32.9 \\ 
        \cmark & \cmark & 0.1 & 0.63 & 0.84 & 11.4 \\ 
        \cmark & \cmark & 1.0 & 0.63* & 0.84* & 7.1* \\ 
        \hline
    \end{tabular}
\end{table}

\begin{figure*}[t]
    \centering
    \includegraphics[width=1.0\linewidth]{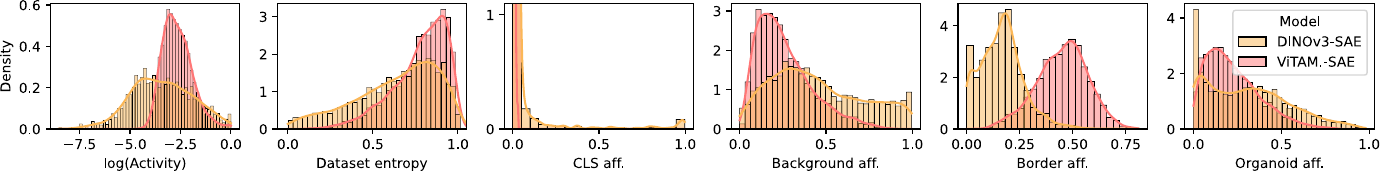}
    \caption{Distribution of concept annotations. DINOv3-SAE has more rarely active and dataset-specific (low entropy) concepts.}
    \label{fig:concept-annotation-distributions}
\end{figure*}

We learn concepts from local and global tokens of the last layer of \vitamino and DINOv3-large. 
\vitamino has a final-layer feature map of $7 \times 7$ tokens of dimension 1536, which are average-pooled for image-level representations~\cite{vitamino}. 
For DINOv3 (large), we obtain $14 \times 14$ patch tokens, a class token and 4 register tokens~\cite{dinov3}. 
We use the patch and class tokens for SAE training.
Across 60652 organoid image crops per epoch, this results in 3\,M tokens for \vitamino and 12\,M tokens for DINOv3. 
We estimate the convex hull $\matr{C}$ of the representation space with bisecting k-means with $k = 32000$ for a dictionary size $c = 4000$ and allow 32 codes per token to be active in BatchTopK. 
As preprocessing, we standardize the tokens per dimension and train for 100 (\vitamino) or 25 (DINOv3) epochs, such that more than 300M tokens were seen per training run. Concept activations are max-pooled across patch and global tokens to create per-image barcodes.

To choose training parameters, we compared the training $R^2$, 
percentage of dead codes, and the learned concept stability.
Concept stability is measured as the cosine similarity of concepts from training runs with different random seeds, where concepts are matched with Hungarian matching~\cite{RA_SAE}. 
Table~\ref{table:ablation-sae-params} shows that the auxiliary top-1 loss reduces dead codes for \vitamino, and replacing $\matr{W}$ with $\operatorname{softmax}(\matr{W})$ increases stability of DINOv3 concepts. Increasing $\delta$ from 0.1 to 1 further reduced dead codes.

For downstream classification, we train logistic regression models and compute the Area Under the Receiver Operating Characteristic Curve (AUROC). Error bars originate from two different SAEs trained with different random seeds.

\subsection{Concept annotation}
\label{sec:interpretable-concept-annotation-organoids:concept-annotation}

To interpret concepts, we annotate several properties: 
\textit{position} (background, border, organoid, and global token-affinity), \textit{dataset specificity}, \textit{activation frequency}, and \textit{task relevance}. 
Further, we retrieve the top 5 activating images as prototypes for a concept and plot the concept activation per patch token, which enables localization and qualitative interpretation. 
All annotations except task relevance use a balanced annotation pool of 1000 organoid crops per source across five datasets (Tellu, OrganoID, OrgaSegment, OrgaQuant, OrgaExtractor). 
Task relevance is evaluated on the training dataset for each task.

For position annotation, we categorize tokens as global (class or average pooled), background, border (patch tokens with both foreground and background pixels), or organoid (foreground). For each concept, we count the prevalence of each token type where the concept is active and estimate the lower bound of a 90 \% confidence interval (CI) for each token type probability. We use a CI lower bound instead of the observed frequency because the sample size where concepts are active varies.
The activation frequency is computed as percentage of images in which a concept is active. The dataset specificity is computed as the entropy of the dataset label on the set of images where a concept is active.  Low entropy means a concept is active mostly in one dataset, while a high entropy indicates a more general concept. To account for different activation frequencies, we normalize the entropy by the mean entropy across random selections of images. 

We compute how relevant a concept is for a single task by computing the adjusted mutual information between the thresholded image barcode activations of a concept and the task labels, taking the highest value across thresholds. Formally, with task classification labels $l \in \{0, ..., n\}^N$ for $N$ training samples, image barcode activations $\vect{i} \in [0, \infty)^N$, the task relevance score is
\begin{align}
    T(\vect{i}, l) = \max_{t\in \{0.0, 0.1, ..., 0.9, 1\}} AMI(l, \mathrm{1}_{\{\vect{i} > t\}}) \, .
\end{align}

\section{Datasets}
\label{sec:datasets}

For SAE training, 
we extract single organoid crops resized to $224 \times 224$ pixels from five public datasets (OrganoID~\cite{OrganoID}, OrgaQuant~\cite{OrgaQuant}, OrgaSegment~\cite{OrgaSegment}, OrgaExtractor~\cite{OrgaExtractor}, Tellu~\cite{Tellu}), which cover diverse organoid types.

We benchmark all models on five diverse downstream tasks described in \vitamino~\cite{vitamino}, containing anterior placode~\cite{vitamino}, otic~\cite{vitamino}, abnormal brain organoids (BO)~\cite{brain_organoids}, budding brain organoids~\cite{brain_organoids}, and ductal organoids (PDLOs)~\cite{vitamino}. 
The anterior placode and otic test sets are in-domain for \vitamino, although the corresponding training data is not used during SAE training. The remaining three test sets are out-of-domain for all methods.

We segment organoids with the Segment Anything Model (SAM)~\cite{SAM} if no segmentation is available and crop around organoids in the otic and BO datasets to reduce the background. However, since we observe cropping-related confounders (visible padding regions) for the BO tasks, we report all main results for BO tasks without cropping.

\begin{figure*}[htb]
    \centering
    \includegraphics[width=1.0\linewidth]{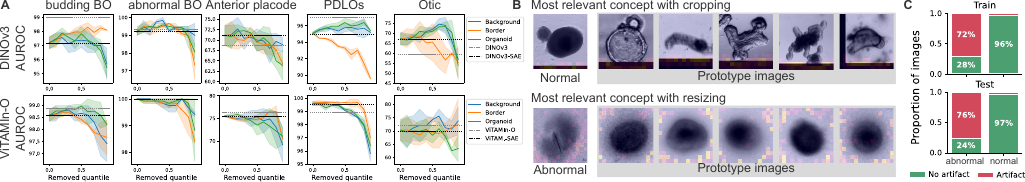}
    \caption{A) Downstream performance when filtering based on concept annotations. B) Prototype images for the most relevant abnormal brain organoid classification concept reveal cropping artifacts. C) Cropping artifacts correlate with the task label.}
    \label{fig:downstream-results}
\end{figure*}

\section{Results}
\label{sec:results}

\subsection{SAEs capture relevant information in few concepts}
\label{sec:results:SAE-concepts}

\begin{figure}[htb]
    \centering
    \centerline{\includegraphics[width=\linewidth]{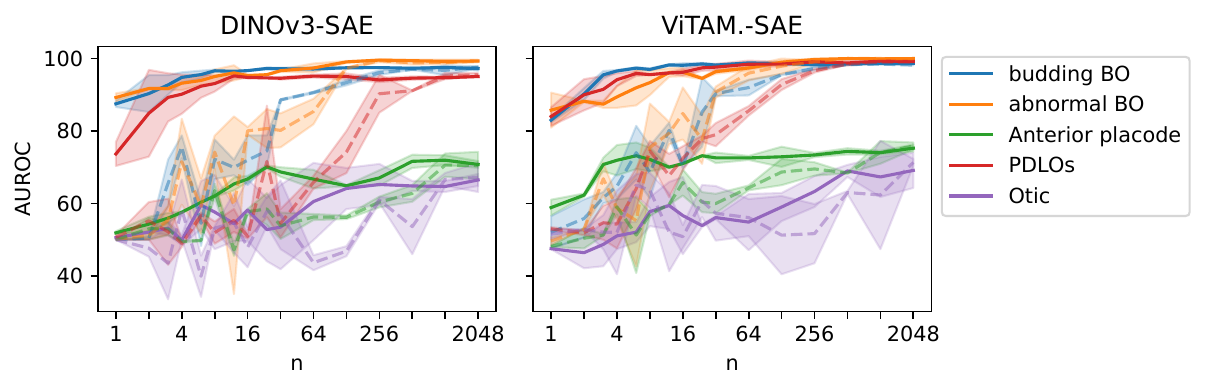}}
    \caption{AUROC using the top-n task-relevant concepts (solid lines, ranked by $T(\vect{i}, l)$ computed on the training split) vs. n randomly selected concepts (dashed lines). 
    }
    \label{fig:concept-task-relevance}
\end{figure}

Classification based on DINOv3 and \vitamino features outperforms CellProfiler features in almost all cases, with \vitamino outperforming DINOv3 (Fig.~\ref{fig:methods-overview}B). DINOv3 and \vitamino-based
SAE features either slightly reduce or even improve performance (for DINOv3 on anterior placode and otic organoids) compared to their non-SAE counterparts, but remain comparable or better than CellProfiler features (Fig.~\ref{fig:methods-overview}B).
Downstream performance with fewer than 32 of the most task-relevant concepts already yields AUROC scores comparable to all concepts for all except the otic organoid task (Fig.~\ref{fig:concept-task-relevance}). 
Top activating prototype images retrieved from the annotation pool for the budding brain organoid task show coherent concepts related to texture near the organoid border or a dark and smooth or blurry organoid border (Fig.~\ref{fig:methods-overview}C).

\subsection{Differences in DINOv3 and \vitamino concepts}
\label{sec:results:dino-vs-vitamino}

\begin{figure}[htb]
    \centering
    \centerline{\includegraphics[width=\linewidth]{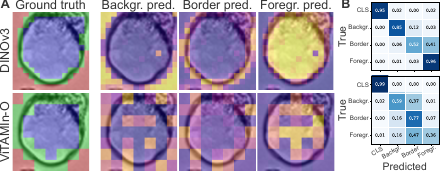}}
    \caption{A) Token-type classification enables segmentation (PDLO example). B) Confusion matrix across all datasets.}
    \label{fig:segmentation-results}
\end{figure}

\vitamino concepts activate more uniformly compared to DINOv3 (Fig.~\ref{fig:concept-annotation-distributions}). 
DINOv3 appears to have more  dataset-specific concepts (low entropy) and more concepts specific to background or foreground. \vitamino has more concepts related to the organoid border. These differences in position annotations could stem from the coarser grid of $7 \times 7$ patch tokens for \vitamino compared to $14 \times 14$ DINOv3 tokens. 

We test whether position annotations transfer to the test datasets. To this end, we compute the average background, border and foreground-affinity per token weighted by the concept activations and classify the token type with logistic regression, trained on the annotation image pool. 
Evaluated across the five test datasets, DINOv3 concepts outperform \vitamino concepts (81.7 vs. 66.4\,\% balanced accuracy). Qualitatively, DINOv3 concepts provide a coarse segmentation of background, organoid border, and foreground, while \vitamino concepts more often misclassify patches (Fig.~\ref{fig:segmentation-results}).

\subsection{Interpretability can improve downstream performance and identify confounders}
\label{sec:results:downstream-interpretability-gains}

Filtering out concepts by removing a quantile of features with the lowest border-affinity improves budding brain organoid classification based on DINOv3-SAE features. Similarly removing DINOv3 concepts with high background or low foreground-affinity improves downstream performance for the PDLO and otic organoid tasks, depending on the quantile threshold (Fig.~\ref{fig:downstream-results}A). \vitamino concepts do not show the same effect, potentially due to the coarse position attribution.

Prototype images for the most task-relevant concept for abnormal brain organoid classification with cropping reveal attention to cropping artifacts (Fig.~\ref{fig:downstream-results}B), which correlate with the classification labels (Fig.~\ref{fig:downstream-results}C). Without cropping during preprocessing, the most relevant concept appears to focus on blurry organoid images with a noisy surrounding texture.

\section{Discussion}
\label{sec:discussion}

We train SAEs to learn interpretable, morphological concepts from two state-of-the-art representation models (DINOv3 and \vitamino), and annotate them with spatial labels, prototype images, dataset specificity and activation frequency. 
We observe that spatial labels from DINOv3 transfer to out-of-domain datasets and enable rough segmentation of organoid, border, and background. Further, prototype images for abnormal brain organoid classification unveil a preprocessing-related confounding factor, which provides an example how interpretable concepts can be used to investigate the reliability of a downstream classifier. 
Future work could further investigate concepts at other layers of the studied representation models, improve concept attribution with patch-level intensity or texture features, or study how to align concepts with biomedical language towards mechanistic reasoning.

\section{Compliance with ethical standards}
\label{sec:ethics}

No ethical approval was required for this study.

\section{Acknowledgments}
\label{sec:acknowledgments}

L. J. G. was supported by the Munich School for Data Science. The authors have no relevant financial or non-financial interests to disclose. Claude was used to improve readability throughout the manuscript, to discuss related work and research ideas, and partially to assist with writing code. All content was written by the authors, and all AI-generated code was reviewed for correctness.

\bibliographystyle{IEEEbib}
\bibliography{IEEEabrv,refs}

@InProceedings{CytoSAE,
    author = { Dasdelen, M.F. and Lim, H. and Buck, M. and Götze, K.S. and Marr, C. and Schneider, S.},
    title = { { CytoSAE: Interpretable Cell Embeddings for Hematology } },
    booktitle = {proceedings of Medical Image Computing and Computer Assisted Intervention -- MICCAI 2025},
    year = {2025},
    publisher = {Springer Nature Switzerland},
    volume = {LNCS 15973},
    month = {September},
    page = {77 -- 86}
}

@article{vitamino,
	author = {Hamurcu, F. and Breunig, M. and Varga, A. and others},
	title = {{ViTAMIn-O: Democratizing computer vision-based machine learning for stem cell research}},
	elocation-id = {2026.06.01.726000},
	year = {2026},
	doi = {10.64898/2026.06.01.726000},
	publisher = {Cold Spring Harbor Laboratory},
	URL = {https://www.biorxiv.org/content/early/2026/06/03/2026.06.01.726000},
	eprint = {https://www.biorxiv.org/content/early/2026/06/03/2026.06.01.726000.full.pdf},
	journal = {bioRxiv}
}

@article{OrganoID,
    doi = {10.1371/journal.pcbi.1010584},
    author = {Matthews, J.M. and Schuster, B. and Kashaf, S.S. and others},
    journal = {PLOS Computational Biology},
    publisher = {Public Library of Science},
    title = {{OrganoID: A versatile deep learning platform for tracking and analysis of single-organoid dynamics}},
    year = {2022},
    month = {11},
    volume = {18},
    url = {https://doi.org/10.1371/journal.pcbi.1010584},
    pages = {1-16},
    number = {11},

}

@article{OrgaSegment,
  title={{OrgaSegment: deep-learning based organoid segmentation to quantify CFTR dependent fluid secretion}},
  author={Lefferts, J.W. and Kroes, S. and Smith, M.B. and others},
  journal={Communications biology},
  volume={7},
  number={1},
  pages={319},
  year={2024},
  publisher={Nature Publishing Group UK, London}
}

@article{Tellu,
    author = {Domènech-Moreno, E. and Brandt, A. and Lemmetyinen, T.T. and Wartiovaara, L. and Mäkelä, T.P. and Ollila, S.},
    title = {{Tellu} – an object-detector algorithm for automatic classification of intestinal organoids},
    journal = {{Disease Models \& Mechanisms}},
    volume = {16},
    number = {3},
    pages = {dmm049756},
    year = {2023},
    month = {03},
    issn = {1754-8403},
    doi = {10.1242/dmm.049756},
}

@article{OrgaQuant,
  title={{OrgaQuant}: human intestinal organoid localization and quantification using deep convolutional neural networks},
  author={Kassis, T. and Hernandez-Gordillo, V. and Langer, R. and Griffith, L.G.},
  journal={Scientific reports},
  volume={9},
  number={1},
  pages={12479},
  year={2019},
  publisher={Nature Publishing Group UK, London}
}

@article{OrgaExtractor,
  title={Development of a deep learning based image processing tool for enhanced organoid analysis},
  author={Park, T. and Kim, T.K. and Han, Y.D. and Kim, K.-A. and Kim, H. and Kim, H.S.},
  journal={Scientific reports},
  volume={13},
  number={1},
  pages={19841},
  year={2023},
  publisher={Nature Publishing Group UK, London}
}

@InProceedings{MorphoOrgaAgent,
        author = { Zhang, H. and Hoermann, M. and Gleiter, L.J. and others},
        title = { { MorphoOrgaAgent: A Foundation-Model-Based Multi-Agent System for Autonomous Organoid Analysis } },
        booktitle = {Medical Image Computing and Computer Assisted Intervention -- MICCAI 2026 Workshops and Challenges},
        year = {2026},
        publisher = {Springer Nature Switzerland},
        volume = {LNCS 17263},
}

@article {DINO4Cells,
	author = {Doron, M. and Moutakanni, T. and Chen, Z.S. and others},
	title = {Unbiased single-cell morphology with self-supervised vision transformers},
	elocation-id = {2023.06.16.545359},
	year = {2023},
	doi = {10.1101/2023.06.16.545359},
	publisher = {Cold Spring Harbor Laboratory},
	URL = {https://www.biorxiv.org/content/early/2023/06/18/2023.06.16.545359},
	eprint = {https://www.biorxiv.org/content/early/2023/06/18/2023.06.16.545359.full.pdf},
	journal = {bioRxiv}
}

@inproceedings{ACE,
 author = {Ghorbani, A. and Wexler, J. and Zou, J. and Kim, B.},
 booktitle = {Advances in Neural Information Processing Systems},
 pages = {},
 publisher = {Curran Associates, Inc.},
 title = {Towards Automatic Concept-based Explanations},
 url = {https://proceedings.neurips.cc/paper_files/paper/2019/file/77d2afcb31f6493e350fca61764efb9a-Paper.pdf},
 volume = {32},
 year = {2019}
}

@inproceedings{into-the-rabbit-hull,
 author = {Fel, T. and Wang, B. and Lepori, M. and others},
 booktitle = {International Conference on Learning Representations},
 pages = {154538--154584},
 title = {{Into the Rabbit Hull: From Task-Relevant Concepts in DINO to Minkowski Geometry}},
 url = {https://proceedings.iclr.cc/paper_files/paper/2026/file/fa68d2dfd8eb6202e7902586197f7629-Paper-Conference.pdf},
 volume = {2026},
 year = {2026}
}

@misc{TopKSAE,
      title={Scaling and evaluating sparse autoencoders}, 
      author={L. Gao and T.D. la Tour and H. Tillman and others},
      year={2024},
      eprint={2406.04093},
      archivePrefix={arXiv},
      primaryClass={cs.LG},
      url={https://arxiv.org/abs/2406.04093}, 
}

@inproceedings{BatchTopKSAE,
 author = {Leask, P. and Bussmann, B. and Pearce, M. and others},
 booktitle = {International Conference on Learning Representations},
 pages = {53617--53642},
 title = {Sparse Autoencoders Do Not Find Canonical Units of Analysis},
 url = {https://proceedings.iclr.cc/paper_files/paper/2025/file/84ca3f2d9d9bfca13f69b48ea63eb4a5-Paper-Conference.pdf},
 volume = {2025},
 year = {2025}
}

@InProceedings{RA_SAE,
  title = 	 {Archetypal {SAE}: Adaptive and Stable Dictionary Learning for Concept Extraction in Large Vision Models},
  author =       {Fel, T. and Lubana, E.S. and Prince, J.S. and others},
  booktitle = 	 {Proceedings of the 42nd International Conference on Machine Learning},
  pages = 	 {16543--16572},
  year = 	 {2025},
  volume = 	 {267},
  month = 	 {Jul},
  publisher =    {PMLR},
  url = 	 {https://proceedings.mlr.press/v267/fel25a.html},
}

@article{dinov3,
  title={{DINOv3}},
  author={Sim{\'e}oni, O. and Vo, H.V. and Seitzer, M. and others},
  year={2025},
  eprint={2508.10104},
  archivePrefix={arXiv},
  primaryClass={cs.CV},
  url={https://arxiv.org/abs/2508.10104},
  journal={arXiv:2508.10104},
}

@article {brain_organoids,
	author = {Kahveci, B. and Polatli, E. and Evranos, A.E. and others},
	title = {{BrAIn: A comprehensive artificial intelligence-based morphology analysis system for brain organoids and neuroscience}},
	elocation-id = {2025.02.19.638973},
	year = {2025},
	doi = {10.1101/2025.02.19.638973},
	publisher = {Cold Spring Harbor Laboratory},
	URL = {https://www.biorxiv.org/content/early/2025/02/24/2025.02.19.638973},
	eprint = {https://www.biorxiv.org/content/early/2025/02/24/2025.02.19.638973.full.pdf},
	journal = {bioRxiv}
}

@article{CellProfiler,
    title={{CellProfiler™: Free, Versatile Software for Automated Biological Image Analysis}},
    author={Lamprecht, M.R. and Sabatini, D.M. and Carpenter, A.E.},
    journal={biotechniques},
    volume={42},
    number={1},
    pages={71--75},
    year={2007},
    publisher={Taylor \& Francis},
    doi = {10.2144/000112257},
    note ={PMID: 17269487},
    URL = {https://doi.org/10.2144/000112257},
}

@article{SAM,
  title={Segment Anything},
  author={Kirillov, A. and Mintun, E. and Ravi, N. and others},
  journal={arXiv:2304.02643},
  year={2023},
}

\end{document}